\documentclass[10pt,twocolumn,letterpaper]{article}

\usepackage{wacv}
\usepackage{times}
\usepackage{epsfig}
\usepackage{graphicx}
\usepackage{amsmath}
\usepackage{amssymb}
\usepackage{booktabs}

\usepackage{xspace}
\usepackage{makecell}
\usepackage{verbatim}
\usepackage[ruled]{algorithm2e}
\usepackage{algorithmic}
\usepackage{defs_ahn_arxiv}
\usepackage{url}

\newtheorem {prop} {Proposition}[section]
\newcommand{\CLEVR}{\textsc{CLEVR}}
\newcommand{\CLEVRTEX}{\textsc{ClevrTex}}
\newcommand{\TEST}{\textsc{OOD}} 
\newcommand{\CAMO}{\textsc{CAMO}}

\usepackage[svgnames]{xcolor}    
\usepackage{siunitx}
\usepackage{multirow}

\DeclareMathOperator{\softargmax}{soft-argmax}
\DeclareMathOperator{\softmax}{softmax}

\wacvfinalcopy 

\ifwacvfinal
\usepackage[breaklinks=true,bookmarks=false]{hyperref}
\else
\usepackage[pagebackref=true,breaklinks=true,colorlinks,bookmarks=false]{hyperref}
\fi

\usepackage[capitalise]{cleveref}
\begin{document}

\title{Unsupervised multi-object segmentation using attention and soft-argmax}

\author{Bruno Sauvalle \quad \quad  Arnaud de La Fortelle\\
    Centre de Robotique\\
  Mines ParisTech PSL University \\
{\tt\small  \{bruno.sauvalle,arnaud.de\_la\_fortelle\}@mines-paristech.fr} }

\maketitle
\thispagestyle{empty}

\begin{abstract}
   We introduce a new architecture for unsupervised object-centric representation learning and multi-object detection and segmentation, which uses a translation-equivariant attention mechanism to predict the coordinates of the objects present in the scene and to associate a feature vector to each object. A transformer encoder handles occlusions and redundant detections, and a convolutional autoencoder is in charge of  background reconstruction. We show that this architecture  significantly outperforms the state of the art on complex synthetic benchmarks.
\end{abstract}


\section{Introduction}

We consider in this paper the tasks of object-centric representation learning and unsupervised object detection and segmentation: Starting from a dataset of images showing various  scenes cluttered with objects, our goal is to build a structured object-centric representation of these scenes, i.e. to map each object present in a scene to a vector representing this object and allowing to recover its appearance and segmentation mask. This task is very challenging because the objects appearing in the images may have different shapes, locations, colors or textures, can occlude each other, and we do not assume that the images share the same background. However the rewards of object-centric representations could be significant since they allow to perform complex reasoning on images or videos \cite{Ding2021,Tang2022} and to learn better policies on downstream tasks involving object manipulation or localization \cite{Veerapaneni2019,Zadaianchuk2020}. The main issue with object-representation learning today is however that existing models are able to process synthetic toy scenes with simple textures and backgrounds but fail to handle more complex or real-world scenes \cite{Laina2021}.

We propose to improve upon this situation by introducing a translation-equivariant and attention-based approach for unsupervised object detection, so that a translation of the input image leads to a similar translation of the coordinates of the detected objects, thanks to an attention map which is used not only to associate a feature vector to each object present in the scene, but also to predict the coordinates of these objects.

The main contributions of this paper are the following: 

\begin {itemize}
\item We propose a theoretical justification for the use of attention maps and soft-argmax for object localization.
\item We introduce a new translation-equivariant and attention-based object detection and segmentation architecture  which does not rely on any spatial prior.
\item We show that the proposed model substantially improves upon the state of the art on unsupervised object segmentation on complex synthetic benchmarks.
\end {itemize}

The paper is organized as follows: In section 2, we provide some theoretical motivation for using attention maps and soft-argmax for object localization. In section 3, we review related work on unsupervised object instance segmentation.  In section 4 we  describe the proposed  model. Experimental results are then provided in section 5.

\section {Motivation for using attention maps  and soft-argmax for object localization}

It is widely recognized that the success of convolutional neural networks is associated with the fact that convolution layers are equivariant with respect to the action of the group of translations, which makes these layers efficient for detecting features which naturally have this property.   It is also easy to show that linear convolution operators are the only linear operators which are equivariant with respect to the natural action of the translation group on feature maps.

We introduce the following notations to describe the action of the translation group: We consider a grayscale image  as a scalar-valued function $\varphi (i,j)$ defined on $\mathbb {Z}^2$ and an element of the group of translations as a vector $(u,v)$ in $\mathbb {Z}^2$. The natural action $T$ of the group of translations on an image can be described by the formula 
\begin{equation} T_{u,v}(\varphi)(i,j) = \varphi (i-u,j-v). \end {equation}
A model layer  $L$  is called equivariant with respect to translations if it satisfies
\begin{equation} L(T_{u,v} \varphi) = T_{u,v}(L(\varphi)). \end {equation}
Let's now consider a localization model $ M$ which takes as input an image  $\varphi(i,j)$  showing one object and produces as output the coordinates of the object present in this image. Such a model does not produce a feature map, so that the previous definition of translation equivariance can not be used for this model.  We remark however that the group of translations acts naturally on $\mathbb {Z}^2$ by the action $T'_{u,v}(i,j) = i+u,j+v$, and that the model $ M$ should have  the equivariance property  
\begin{equation} M(T_{u,v} \varphi) = T'_{u,v}(M(\varphi)).\label{equi2} \end{equation}
Indeed, if the complete image is translated by a vector $(u,v)$, then the object present in this image is also translated, so that the associated coordinates have to be shifted according to the vector $(u,v)$.

It is not difficult to see that  in the same way that convolutional operators are the only linear operators equivariant with respect to translations, it is also possible to fully  describe which elementary operators follow this specific equivariance property. We first remark however that we have to restrict the space of possible input maps $\varphi$: if $\varphi$ is a constant function, it does not change under the action of the translation group, so that the equivariance property \ref{equi2} cannot be satisfied with such a function. We then suppose that $\varphi$ satisfies $\sum_p \varphi(p) = 1$ and consider that the domain of the operator $M$  is the corresponding affine space $\mathcal A$. We also replace the linearity condition by an the following affinity condition: 

For all $\alpha_i \in \mathbb R, \varphi_i \in \mathcal A$ so that $\sum_i \alpha_i = 1$, we have $ M(\sum_i \alpha_i \varphi_i) = \sum_i \alpha_i  M(\varphi_i) $.

We then have the following proposition: 
\begin{prop} \label{prop1}

An affine operator $ M$ which satisfies the equivariance property \ref{equi2}  has to be of the form 
\begin{equation} M(\varphi) = C+ \sum_{p \in \mathbb Z^2} \varphi(p)p \end{equation}
for some constant C in $\mathbb R^2$.
\end {prop}
Proof: We  write  the input map $\varphi$ as a sum of spatially shifted version of the function $\delta \in \mathcal A$ satisfying $\delta(p) = 1$ for $p=(0,0)$ and $\delta(p) = 0$ for $p \neq (0,0)$: 
 \begin {equation} \varphi(p) = \sum_{q \in \mathbb {Z}^2}\varphi(q )\delta(p-q) \end{equation}
 
  We then  use the the  affine property of $ M$ and equivariance property \ref{equi2}:
 
 \begin{equation}   M(\varphi) =  M ( \sum_q\varphi(q)\delta(p-q)) \end{equation}
 \begin{equation} = \sum_q\varphi(q) M( \delta(p-q)) = \sum_q\varphi(q)( M( \delta) +q)  \end{equation}
 \begin{equation} = (\sum_q\varphi(q)) M(\delta)+\sum_q  \varphi(q)q \end{equation}
 \begin{equation} =  M( \delta)+\sum_q  \varphi(q)q, \end {equation}
 which proves the proposition since $ M( \delta)$ is a constant.

The proposition \ref{prop1} can be interpreted as stating that in order to get an equivariant localization operator, the most straightforward method is to  build a normalized attention map $\varphi$ from the input image and compute the coordinates of the detected object using an attention mechanism with $\varphi$ as attention map and pixel coordinates as target values. 
One remarks that it is precisely what the soft-argmax operator is doing: It takes an unnormalized scalar map $\phi$ as input, normalizes it using a softmax operator, and then perform localization using the same formula as in $ \ref{prop1}$: 

\begin{equation}\begin{split} \softargmax(\phi) & = \sum_{p \in \mathbb Z^2} \softmax(\phi)(p) p  \\ & =  \sum_{p \in \mathbb Z^2} \frac {e^{\phi(p)}} {\sum_{q \in \mathbb Z^2} e^{\phi(q)}} p \end{split} \end{equation}

This operation is called soft-argmax because it allows to compute in a differentiable way an estimate of the coordinates of the maximum of the input map $\phi$. Using soft-argmax then appears to be the most natural way to get an equivariant localization operator.

\section{Related work}

\paragraph {Unsupervised object detection and segmentation} Unsupervised object detection and segmentation models are generally reconstruction models: They try to reconstruct  the input image using a specific image rendering process which induces the required object-centric structure. In order to ensure that objects are properly detected, various objectness priors have  been defined and implemented: 
\begin {itemize}
\item pixel similarity priors. Some models consider the task of object segmentation as a clustering problem, which can be addressed using deterministic \cite{Hwang2019,Locatello2020a} or probabilistic  \cite{Engelcke2021, Greff2016, VanSteenkiste2018} methods: If the feature vectors associated to two different pixels of an image are very similar, then it is considered that these pixels should both belong  to the same object or to the background.

\item  independence priors. Some models assume that the images are sampled from a distribution  which follows a probabilistic model featuring some independence priors between objects and the background,  and use variational  \cite{Greff2019a, Engelcke2019} or adversarial \cite{Chen,Bielski2019} methods to learn these distributions.

\item disentanglement of appearance and location.  Foreground objects appearing in the scenes of a given dataset can have similar shapes and appearances but very different scales and locations. Object discovery is performed  by disentangling the object appearance generation process, which is performed by a convolutional glimpse generator  \cite{AliEslami2016,Kosiorek2018a,Crawford2019,Stelzner2019,Jiang,Jiang2020} or a learned dictionary \cite{Monnier2021,Smirnov2021}, from the translation and scaling of the objects appearing in a scene, which is usually done by including a spatial transformer network  \cite{Jaderberg2015a} in the model. The model described in this paper belongs to this category and uses an convolutional glimpse generator.

\end {itemize}

\paragraph{Object detection and segmentation without spatial prior} State of the art supervised detection and segmentation models usually rely on predefined reference anchors or center points which are spatially organized according to a periodic grid structure. The use of periodic grids has also been proposed  for unsupervised object detection \cite{Lin2020,Jiang,Jiang2020,Smirnov2021}. Alternative  detection methods relying on heatmaps produced by a U-net  \cite{10.1007/978-3-319-24574-4_28} or stacked U-nets   \cite{Newell2016} networks, which predict for each pixel the probability of presence of one object on this pixel have been implemented in the supervised setting \cite{Law2020,Duan2019a}. 

For some specific applications such as human pose estimation   or anatomical landmark localization \cite{Tiulpin}, some supervised models predict one heatmap per object. The use of soft-argmax for converting heatmaps to object coordinates has been implemented in the supervised \cite{Sun2018a,Luvizon2019a, Chandran2020},  semi-supervised    \cite{Honari2018} and unsupervised settings \cite{Goroshin2015,Finn2016} but has never been proposed for unsupervised object detection or segmentation.

More recently, transformer-based \cite{Vaswani2017a} models using object \cite{Carion2020,Zhu2020,Dong2021} or mask \cite{Cheng2021, Cheng2021a} queries have been proposed which not not rely explicitly on a spatial grid. These models show that transformers are efficient in the supervised setting to avoid multiple detections of the same object. 

\section{Description of proposed model }

\begin{figure*}
 \vspace{-5mm}
\centering   
 \includegraphics[width=14cm]{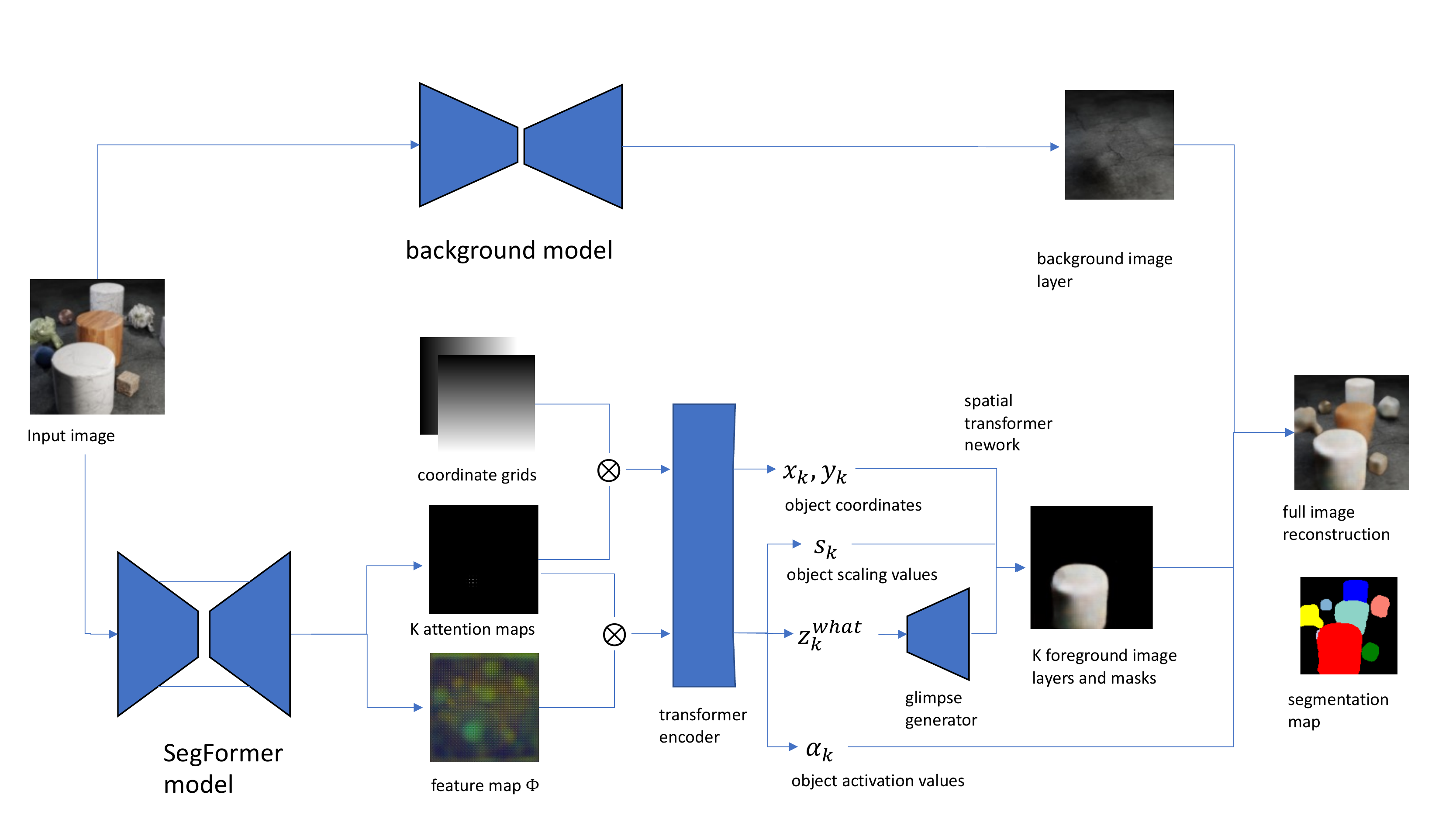} 
\caption{\textbf {Overview of proposed model}.  A High resolution feature map generator (Segformer model) is trained to produce a high resolution feature map $\Phi$ and K scalar attention maps (one per object query). These maps are used to predict the coordinates and scales of the detected objects and the associated feature vectors, which are refined by a transformer encoder and then used as inputs to a glimpse generator and a spatial transformer network to produce K object image layers and masks. A convolutional autoencoder  is in charge of  background reconstruction.}
\label{fig:model_diagram}
 \vspace{-5mm}
\end{figure*}
\subsection{Model architecture}

The overall architecture of the model is described in Fig \ref{fig:model_diagram}.

The proposed model is composed of a  a foreground model and a background model.

The  background model is a deterministic convolutional autoencoder: We rely on the classical assumption \cite{Wright2009} that background images lie on a low-dimensional manifold, and use the autoencoder to learn this manifold.

The  foreground model is also deterministic and associates to each object in the scene an appearance vector $z^{what}$ which is used to produce a glimpse of the object, which is then scaled and translated at the right position on the image using a spatial transformer network.

The foreground encoding and reconstruction process can be described as follows: First, a high resolution feature map generator takes a color image of size $h \times w$ as input and produces  a high resolution feature map $\Phi$ of dimension $d_\Phi$  and several scalar attention logit maps $A_1,..., A_K$. 
We will use in this paper the transformer-based Segformer model \cite{Xie2021}, which produces feature maps of size $h^* \times w^* = h/4 \times w/4$. The hyperparameter $K$ is set to the maximum number of objects on a scene in the dataset.  The scalar attention logit maps $A_1,..., A_K$ are transformed into a normalized attention maps $\mathcal A_1,..., \mathcal A_K$ using a softmax operator:

\begin{equation}  \mathcal A_k(i,j) = \frac {e^{A_k(i,j)}} {\sum_{i',j'} e^{A_k(i',j')}}  \end {equation}

We normalize the pixel indices $(i,j)$ from the range $[1,..,w^*]$ and $[1,..,h^*]$ to the range $[-1,1]$ required by spatial transformer networks using the formulas 
\begin{equation} x(i) = 2\frac {i-1}{w^*-1} - 1 \end{equation}
\begin{equation} y(j) = 2\frac {j-1}{h^*-1} - 1, \end{equation}
and predict initial estimates $x_k^0,y_k^0$ of the  coordinates of the  detected objects as the  the center of mass of the attention maps $\mathcal A_k$: 
\begin{equation}\label{eq:argmax1}  x_k^0 = \sum_{i=1,j=1}^{w^*,h^*}\mathcal A_k(i,j)x(i) \end{equation}
\begin{equation}\label{eq:argmax2}  y_k^0 = \sum_{i=1,j=1}^{w^*,h^*} \mathcal A_k(i,j)y(j) \end{equation}

 We also build $K$ object query feature vectors $\phi_1^0,...,\phi_K^0$ of dimension $d_\Phi$ using the same attention maps $\mathcal A_1,..,\mathcal A_K$ as weights and the feature map $\Phi$ as target values:  

\begin{equation}\label{eq:aggreg} \phi_k^0 = \sum_{i=1,j=1}^{w^*,h^*}\mathcal A_k(i,j) \Phi(i,j) \end{equation}

A transformer encoder then takes the  K triplets $(\phi_k^0, x_k^0,  y_k^0)_{1 \le k \le K}$  as inputs and produces a refined version $(\phi_k,  x_k,   y_k)_{1 \le k \le K}$  taking into  account possible detection redundancies and object occlusions.
More precisely, we use a learned linear embedding to increase the dimension of  the triplets $(\phi_k^0, x_k^0,  y_k^0)$ from 
$d_{\Phi}+2$ to the input dimension $d_{T}$ of the transformer encoder, and a learned linear projection to reduce the dimension of the outputs of the transformer encoder from $d_{T}$ back to $d_{\Phi}+2$.
 The transformer encoder does not take any positional encoding as input, considering that the transformation which has to be performed should not depend on the ordering of the detections.

 We force the final values of $x_k$ and $y_k$ to stay in the range $[-1,1]$ using clamping.
Each transformed feature vector $\phi_k$ is then split in three terms: $\phi_k = (s_k,  \alpha_k, z_k^{what})$.  
\begin{itemize}
\item The first term $s_k$ is an inverse scaling factor. It is a scalar if objects in the dataset have widths and heights which are similar (isotropic scaling), or a pair of scalars $s^x_k, x_k^y$ if this is not the case (anisotropic scaling).
We force the values of $s_k$  to stay within a fixed range using a sigmoid function. The maximum value of this range ensures that a non-zero gradient will be available. The minimum value is set higher than 1 to make sure that the glimpse generator will not try to generate a full image layer.

 \item The second term is a scalar which is  assumed to predict the activation level $\alpha_k$ of the object, which will be used to predict whether it is visible or not. We force this  activation value to be positive using an exponential map.
 
 \item The remaining coordinates form a vector $z_k^{what}$ which codes for the appearance of the object.

 \end{itemize}
 
We then use a convolutional glimpse generator to build a color image $o_k$ of the associated object together with the associated scalar mask $m_k$, using $z^{what}_k$ as input.
These images and masks are translated to  the positions $(x_k,y_k)$ and scaled according to the inverse scaling factor $s_k$ using a spatial transformer network. We note $ L_k$ and $M_k$ for $k \in \{1,..,K\}$  the corresponding object image layers and masks, and $L_0$ the background image produced by the background model, so that we have a total of $K+1$ image layers. 

We now have to decide for each pixel whether this pixel should show the background layer or one of the $K$ object layers.
In order to do this in a differentiable way, we multiply the predicted object masks $M_k$ with the associated object activation levels $\alpha_k$, and normalize the results to get one normalized weights distribution $(w_k)_{0 \le k \le K}$ per pixel:
\begin{equation} \label{eq:weight}w_k(i,j) =  \frac {\alpha_k M_k(i,j )} { \sum_{k' \in {0..K}}\alpha_{k'} M_{k'}(i,j)},  \end{equation}
considering that the mask $M_0$ associated to the background is set to 1 everywhere and that it has a fixed learned activation factor $\alpha_0$.

The final reconstructed image $\hat X$  is then equal to the weighted sum of the various image layers using the weights $ w_k$: 
\begin{equation} \hat X(i,j) = \sum_{k=0}^K w_k(i,j) L_k(i,j) \end{equation}
During inference the segmentation map is built by assigning to  each pixel the layer index $k \in \{0,..,K\}$ for which $w_k(i,j)$ is the maximum. The background model is not needed to get the segmentation maps during inference.

\subsection{Model training}

\subsubsection{loss function}

In order to train the proposed model, we use a main reconstruction loss function and an auxiliary loss:
\paragraph {reconstruction loss}

The local $L_1$ reconstruction error associated to the pixel $(i,j)$ is  
\begin{equation} l_{i,j} = \sum_{c=1}^{3} \lvert \hat x_{c,i,j} -x_{c,i,j} \rvert \label{equ_1}, \end{equation}
where  $x_{c,i,j}$ and $\hat x_{c,i,j}$ are the values of  the color channel $c$  at the position $(i,j)$ in the input image and reconstructed image.

The reconstruction loss is defined as the mean square of this reconstruction error. 
\begin {equation}  \mathcal L_{rec} = \frac {1} {hw} \sum_{i=1,j=1}^{w,h} l_{i,j} ^2 \end {equation}

\paragraph {pixel entropy loss} For a given pixel $(i,j)$, we expect the distribution of the weights $w_0(i,j), .., w_K(i,j)$ to be one-hot, because we assume that the objects are opaque. We observe that a discrete distribution is one-hot if and only if it has a  zero entropy, so that minimizing the entropy of this distribution would be a reasonable way to enforce a stick-breaking process. Considering however that the entropy function has a singular gradient near one-hot distributions, we use the square of the entropy function to build the loss function. We then define the pixel entropy loss as 
\begin {equation}  \mathcal L_{pixel} = \frac {1}{hw} \sum_{i=1,j=1}^{w,h} ( \sum_{k=0}^K  w_{k}(i,j) \log(w_k(i,j)+\epsilon))^2, \end{equation} 

where $\epsilon = 10^{-20}$ is introduced to avoid any numerical issue with the logarithm function.

This auxiliary loss is weighted using the weight $\lambda_{pixel}$ before being added to the reconstruction loss.

During our experiments, we observed that   the pixel entropy loss could prevent a successful initialization of the localization process during the beginning of the training. As a consequence, we smoothly activate this auxiliary loss during initialization using a quadratic warmup of the weight. 

 The full loss function is then equal to 
\begin {equation} \mathcal L = \mathcal L_{rec}+ \min(1, \frac {step}{N_{pixel}})^2\lambda_{pixel} \mathcal L_{pixel}, \end {equation} 
where $step$ is the current training iteration index and $N_{pixel}$ is a fixed hyperparameter.

\subsubsection {curriculum training}

The interaction between the background reconstruction model and the foreground model  during training  is a very challenging issue, because of the competition between them to reconstruct the image. We handle this problem as in \cite{Jiang2020} by implementing curriculum training. We will then evaluate two methods to train the proposed model:
\begin{itemize}
\item baseline training (BT) : The background and foreground models are initialized randomly and trained simultaneously.

\item curriculum training (CT): The training of the model is split in three phases : 
\begin{enumerate}
\item The background model is pretrained alone,  using the methodology and robust loss function described in \cite{Sauvalle2021b}. 
\item The weights of the background model are then frozen and the foreground model is trained using the frozen background model.
\item The background and foreground models are then fine-tuned simultaneously.
\end{enumerate}

\end{itemize}

\section {Experimental results}

\subsection{Evaluation on public benchmarks}

We perform a quantitative evaluation of the proposed model on the following datasets: CLEVRTEX \cite{Laina2021}, CLEVR \cite{Johnson2017}, ShapeStacks \cite{Groth2018}  and ObjectsRoom \cite{multiobjectdatasets19}. 

We implement on ShapeStacks, ObjectsRoom and CLEVR the same preprocessing as in \cite{Engelcke2021}.

 We use the same hyperparameter values on these datasets, except for the hyperparameter $K$ related to the number of object queries, which is set to the maximum number of objects in each dataset (i.e. 3 on ObjectsRoom, 6 on ShapeStacks and 10 on CLEVRTEX and CLEVR). We use isotropic scaling on CLEVR and ShapeStacks and anisotropic scaling on the other datasets.

 We use the  versions B3 of the Segformer model, and rely on the Hugging Face implementation of this model, with pretrained weights on ImageNet-1k for the hierarchical transformer backbone, but random initialization for the MLP decoder which is used as a feature map generator. We use the standard Pytorch implementation of the transformer encoder.  The  architecture of the backgroud model autoencoder is the same as in \cite{Sauvalle2021b}. The  glimpse generator is a sequence of transpose convolution layers, group normalization \cite{Wu2020} layers and CELU \cite{Barron2017} non-linearities, and is described in the supplementary material.

We use Adam as optimizer. The training process includes a quadratic warmup of the learning rate since the model contains a transformer encoder. We also decrease the learning rate by a factor of 10 when the number of training steps reaches 90\% of the total number of training steps. The total number of training steps of the baseline training (BT) scenario is 125 000. In the curriculum training (CT) scenario, the number of training steps for background model pretraining (phase 1) is 500 000 on CLEVRTEX, ShapeStacks and ObjectsRoom, but 2500 on CLEVR, which shows a fixed background, as recommended in \cite{Sauvalle2021b}. The number of training steps of  phase 2 (training with frozen pretrained background model) is 30 000, and the  number of training steps of the final fine-tuning phase (phase 3) is 95 000.

Full implementation details and hyperparameter values are provided in the supplementary material, and the model code will be made available on the Github platform.

In order to compare our results with published models, we compute the following evaluation metrics: mean intersection over union (mIoU) and adjusted rand index restricted to foreground objects (ARI-FG). We also provide  the mean square error (MSE) between the reconstructed image and the input image, which provides an estimate of the accuracy of the learnt representation. We use the same definitions and methodology as \cite{Laina2021} for these metrics. We provide the mean segmentation covering (defined in \cite{Engelcke2019}) restricted to foreground objects (MSC-FG)  on ObjectsRoom and ShapeStacks where mIoU baseline values are not available.

We call AST-Seg (Attention and Soft-argmax with Transformer using Segformer) the proposed model, and AST-Seg-B3-BT, AST-Seg-B3-CT respectively  the models using a Segformer B3 feature map generator trained under the baseline training  or curriculum training scenarios.
 Table \ref{tab:results_cle_clt_aris} and  \ref{tab:shapestacks_objectroom} provide the results obtained on these datasets  with a comparison with published results. 

The proposed model trained under the baseline training scenario gets better average results than existing models on the CLEVR and CLEVRTEX dataset, but shows a very high variance. For example, on the CLEVR dataset,  the model may fall during training in a bad minimum where the background model tries to predict the foreground objects.  Using curriculum training allows to avoid this issue, get stable results on all datasets, and obtain a very significant mIoU improvement on the most complex datasets CLEVR and CLEVRTEX.

\sisetup{mode = text}

\newcommand{\integer}[2]{
\hspace*{-.1em}\tablenum[table-format = 2]{#1}\small\color{DarkSlateBlue}$\pm$\tablenum[table-format = 2]{#2}
}
\newcommand{\boldinteger}[2]{
\hspace*{-.1em}\bfseries\tablenum[table-format = 2]{#1}\small\color{DarkSlateBlue}$\pm$\tablenum[table-format = 2]{#2}
}

\newcommand{\arif}[2]{
\hspace*{-.1em}\tablenum[table-format = 2.2]{#1}\small\color{DarkSlateBlue}$\pm$\tablenum[table-format = 2.2]{#2}
}
\newcommand{\boldarif}[2]{
\hspace*{-.1em}\bfseries\tablenum[table-format = 2.2]{#1}\small\color{DarkSlateBlue}$\pm$\tablenum[table-format = 2.2]{#2}
}
\newcommand{\arifp}[1]{
\hspace*{-.5em}\tablenum[table-format = 2.2]{#1}\hspace*{3.1em}
}

\newcommand{\mious}[2]{
\hspace*{-1.2em}\tablenum[table-format = 2.2]{#1}\small\color{DarkSlateBlue}$\pm$\tablenum[table-format = 2.2]{#2}\hspace*{-1.3em}
}
\newcommand{\boldmious}[2]{
\hspace*{-1.2em}\bfseries\tablenum[table-format = 2.2]{#1}\small\color{DarkSlateBlue}$\pm$\tablenum[table-format = 2.2]{#2}\hspace*{-1.3em}
}

\newcommand{\mioups}[1]{
\hspace*{-1.2em}\tablenum[table-format = 2.2]{#1}\hspace*{2.4em}
}


\newcommand{\mse}[2]{
\tablenum[table-format = 2.0]{#1}\small\color{DarkSlateBlue}$\pm$\tablenum[table-format = 3.0]{#2}\hspace{-1em}
}
\newcommand{\boldmse}[2]{
\bfseries\tablenum[table-format = 2.0]{#1}\small\color{DarkSlateBlue}$\pm$\tablenum[table-format = 3.0]{#2}\hspace{-1em}
}
\newcommand{\msep}[1]{
\tablenum[table-format = 2.0]{#1}\hspace*{1.3em}
}

\begin{table*}
  \centering
\caption{Benchmark results on \CLEVR{} and  \CLEVRTEX{}. Results are shown $(\pm \sigma)$ calculated over 3 runs. Source: \cite{Laina2021}}
\label{tab:results_cle_clt_aris}
\scalebox{0.8}{
\begin{tabular}{lrrrlrrrrr}
\toprule
\multirow{2}{*}{Model} & \multicolumn{3}{c}{\CLEVR{}} 
& 
      & \multicolumn{3}{c}{\CLEVRTEX{}} 

\\
& \(\uparrow\)mIoU (\%) 
& \(\uparrow\)ARI-FG (\%) 
& \(\downarrow\)MSE 
& 
& \(\uparrow\)mIoU (\%) 
& \(\uparrow\)ARI-FG (\%) 
& \(\downarrow\) MSE 
\\
\cmidrule[\lightrulewidth]{1-4}
\cmidrule[\lightrulewidth]{6-8}

 SPAIR \cite{Crawford2019}
& \mious{65.95}{4.02} & \arif{77.13}{1.92} 
&\mse{55}{10} 
&
&\mious{0.00}{0.00} & \arif{0.00}{0.00} 
&\mse{1101}{2} 
\\

SPACE \cite{Lin2020} 
& \mious{26.31}{12.93} & \arif{22.75}{14.04} 
&\mse{63}{3} 

&
& \mious{9.14}{3.46} & \arif{17.53}{4.13} 
&\mse{298}{80} 
\\

 GNM \cite{Jiang2020} 
& \mious{59.92}{3.72}& \arif{65.05}{4.19} 
&\mse{43}{3} 

&
& \mious{42.25}{0.18} & \arif{53.37}{0.67} 
&\mse{383}{2} 
&
\\

MN \cite{Smirnov2021} 
 & \mious{56.81}{0.40} & \arif{72.12}{0.64}
&\mse{75}{1} 
&
& \mious{10.46}{0.10} & \arif{38.31}{0.70} 
&\mse{335}{1} 

\\

 DTI \cite{Monnier2021} 
 & \mious{48.74}{2.17}& \arif{89.54}{1.44}
&\mse{77}{12} 
&
& \mious{33.79}{1.30}& \arif{79.90}{1.37} 
&\mse{438}{22} 
\\

 Gen-V2 \cite{Engelcke2021} 
& \mious{9.48}{0.55} & \arif{57.90}{20.38} 
&\mse{158}{2} 
&
& \mious{7.93}{1.53} & \arif{31.19}{12.41} 
&\mse{315}{106} 

\\

 eMORL \cite{Emami2021} 

& \mious{50.19}{22.56}& \arif{93.25}{3.24} 
&\mse{33}{8} 
&

& \mious{12.58}{2.39}& \arif{45.00}{7.77} 
&\mse{318}{43} 

\\

MONet \cite{Burgess2019} 
& \mious{30.66}{14.87}& \arif{54.47}{11.41} 
&\mse{58}{12} 
&
& \mious{19.78}{1.02}& \arif{36.66}{0.87} 
&\mse{146}{7} 

\\

 SA \cite{Locatello2020a} 
 & \mious{36.61}{24.83} & \arif{95.89}{2.37}
&\mse{23}{3} 
&
& \mious{22.58}{2.07} & \arif{62.40}{2.23} 

&\mse{254}{8} 

\\

 IODINE \cite{Greff2019a} 

& \mious{45.14}{17.85}& \arif{93.81}{0.76} 
&\mse{44}{9} 
&

& \mious{29.17}{0.75}& \arif{59.52}{2.20} 
&\mse{340}{3} 

\\
\cmidrule[\lightrulewidth]{1-8}
 AST-Seg-B3-BT

& \mious{71.92}{32.94}& \arif{76.05}{36.13} 
&\mse{51}{63} 

&

& \mious{57.30}{15.72}& \arif{71.79}{22.88} 
&\mse{152}{39}

\\
 AST-Seg-B3-CT

& \boldmious{90.27}{0.20}& \boldarif{98.26}{0.07} 
&\boldmse{16}{1}

&
& \boldmious{79.58}{0.54}& \boldarif{94.77}{0.51} 
&\boldmse{139}{7}

\\

\bottomrule
\end{tabular}
}
\end{table*}

\begin{table*}[h]
  \centering
\caption{Benchmark results on ObjectsRoom and ShapeStacks. Source: \cite{Engelcke2021}.}
\label{tab:shapestacks_objectroom}
\scalebox{.7}{
\begin{tabular}{lrrcclrrcclrrlrr}
\toprule
\multirow{2}{*}{Model} & \multicolumn{4}{c}{ObjectsRoom{}} 
& 
      & \multicolumn{4}{c}{ShapeStacks} 
\\
& \(\uparrow\)ARI-FG (\%) 
& \(\uparrow\)MSC-FG (\%) 
& \(\uparrow\)mIoU (\%) 
& \(\downarrow\)MSE
& 
& \(\uparrow\)ARI-FG (\%) 
& \(\uparrow\)MSC-FG (\%) 
& \(\uparrow\)mIoU (\%) 
& \(\downarrow\)MSE 
\\
\cmidrule[\lightrulewidth]{1-5}
\cmidrule[\lightrulewidth]{7-10}

MONet-g  \cite{Burgess2019, Engelcke2021} 
& \integer{54}{0} & \integer{33}{1} &  n/a& n/a
&
& \integer{70}{4} & \integer{57}{12} &  n/a& n/a 
\\
Gen-v2 \cite{Engelcke2021}
& \integer{84}{1} & \integer{58}{3} & n/a & n/a 
&
& \boldinteger{81}{0} & \integer{68}{1} & n/a & n/a 
\\

SA \cite{Locatello2020a}
& \integer{79}{2} & \integer{64}{13} & n/a & n/a 
&
& \integer{76}{1} & \integer{70}{5} & n/a & n/a 
\\
\cmidrule[\lightrulewidth]{1-5}
\cmidrule[\lightrulewidth]{7-10}
AST-Seg-B3-BT 
& \arif{74.96}{10.02} & \arif{69.86}{10.13} & \arif{74.50}{8.61} & \arif{11.7}{2.1}
&
& \arif{73.77}{7.56} & \arif{74.12}{8.63} & \arif{70.18}{12.68} & \arif{11.8}{7.0}
%

%

\\
AST-Seg-B3-CT
& \boldarif{87.23}{0.88} & \boldarif{82.22}{0.96} & \boldarif{85.02}{0.79} & \boldarif{6.7}{0.9}
&
& \arif{79.34}{0.73} & \boldarif{77.65}{1.3} & \boldarif{78.84}{0.21} & \boldarif{4.5}{0.2}

\\

\bottomrule
\end{tabular}
}
\end{table*}

Following the methodology proposed in   \cite{Laina2021}, we also evaluated the generalization capability of a model trained on CLEVRTEX when applied to datasets containing out of distribution images showing unseen textures and shapes or camouflaged objects (OOD and CAMO datasets \cite{Laina2021}). The results of this evaluation are provided in Table \ref{tab:results_ood_camo} and show that the proposed model  generalizes well, although it is deterministic and does not use any specific regularization scheme.

\begin{table*}[h]
  \centering
\caption{Benchmark generalization results on \CAMO{}, and \TEST{} for a model trained on CLEVRTEX. Results are shown $(\pm \sigma)$ calculated over 3 runs. Source: \cite{Laina2021}}

\label{tab:results_ood_camo}
\scalebox{0.8}{
\begin{tabular}{lrrrlrrrrr}
\toprule
\multirow{2}{*}{Model} & \multicolumn{3}{c}{\TEST{}} 
& 
      & \multicolumn{3}{c}{\CAMO{}} 

\\
& \(\uparrow\)mIoU (\%) 
& \(\uparrow\)ARI-FG (\%) 
& \(\downarrow\)MSE 
& 
& \(\uparrow\)mIoU (\%) 
& \(\uparrow\)ARI-FG (\%) 
& \(\downarrow\)MSE 
\\
\cmidrule[\lightrulewidth]{1-4}
\cmidrule[\lightrulewidth]{6-8}

 SPAIR \cite{Crawford2019}
& \mious{0.00}{0.00} & \arif{0.00}{0.00} 
&\mse{1166}{5} 
&
& \mious{0.00}{0.00} & \arif{0.00}{0.00} 
&\mse{668}{3} 
\\

SPACE \cite{Lin2020} 
& \mious{6.87}{3.32} & \arif{12.71}{3.44} 
&\mse{387}{66} 

&
& \mious{8.67}{3.50} & \arif{10.55}{2.09} 
&\mse{251}{61} 
\\

 GNM \cite{Jiang2020} 
& \mious{40.84}{0.30}& \arif{48.43}{0.86} 
&\mse{626}{5} 

&
& \mious{17.56}{0.74} & \arif{15.73}{0.89} 
&\mse{353}{1} 
&
\\

MN \cite{Smirnov2021} 
 & \mious{12.13}{0.19} & \arif{37.29}{1.04}
&\mse{409}{3} 
&
& \mious{8.79}{0.15} & \arif{31.52}{0.87} 
&\mse{265}{1} 

\\

 DTI \cite{Monnier2021} 
& \mious{32.55}{1.08}& \arif{73.67}{0.98} 
&\mse{590}{4} 
&
 & \mious{27.54}{1.55}& \arif{72.90}{1.89}

&\mse{377}{17} 
\\

 Gen-V2 \cite{Engelcke2021} 
& \mious{8.74}{1.64} & \arif{29.04}{11.23} 
&\mse{539}{147} 
&
& \mious{7.49}{1.67} & \arif{29.60}{12.84} 
&\mse{278}{75} 

\\

 eMORL \cite{Emami2021} 

& \mious{13.17}{2.58}& \arif{43.13}{9.28} 
&\mse{471}{51} 
&

& \mious{11.56}{2.09}& \arif{42.34}{7.19} 
&\mse{269}{31} 

\\

MONet \cite{Burgess2019} 
& \mious{19.30}{0.37}& \arif{32.97}{1.00} 
&\boldmse{231}{7} 
&
& \mious{10.52}{0.38}& \arif{12.44}{0.73} 
&\boldmse{112}{7} 

\\

 SA \cite{Locatello2020a} 
 & \mious{20.98}{1.59} & \arif{58.45}{1.87}
&\mse{487}{16} 
&
& \mious{19.83}{1.41} & \arif{57.54}{1.01} 

&\mse{215}{7} 

\\

 IODINE \cite{Greff2019a} 

& \mious{26.28}{0.85}& \arif{53.20}{2.55} 
&\mse{504}{3} 
&

& \mious{17.52}{0.75}& \arif{36.31}{2.57} 
&\mse{315}{3} 

\\
\cmidrule[\lightrulewidth]{1-8}

 AST-Seg-B3-CT

& \boldmious{67.50}{0.75}& \boldarif{83.14}{0.75} 
&\mse{832}{24} 
 
&

& \boldmious{73.07}{0.65}& \boldarif{87.27}{3.78} 
&\mse{145}{6}

\\

\bottomrule
\end{tabular}
}
\end{table*}

\begin{figure*}[h!]

\centering
 \includegraphics[width=18cm]{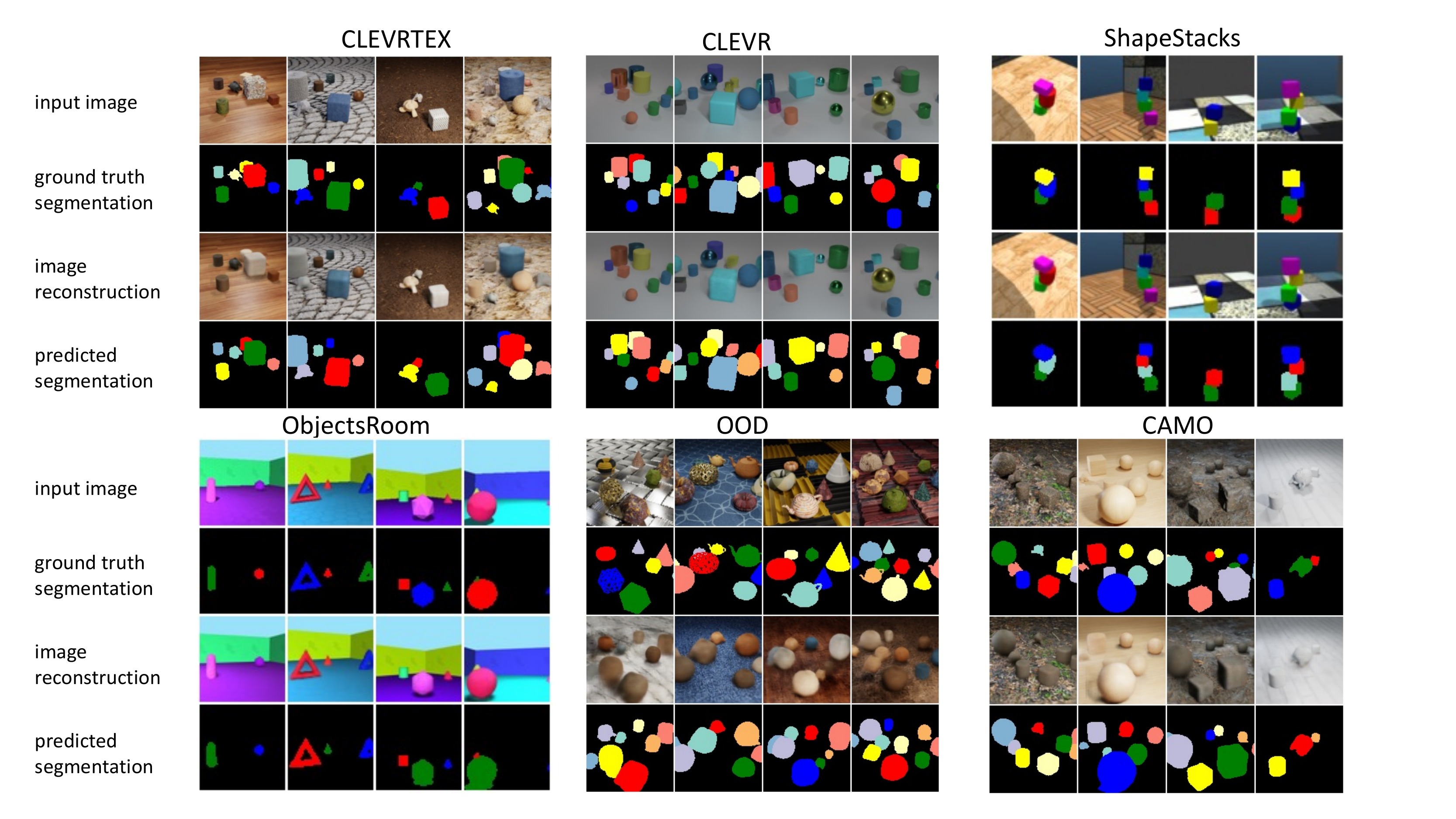} 
\caption{Examples of segmentation predictions on CLEVRTEX, CLEVR, ShapeStacks, ObjectsRoom, OOD and CAMO test datasets (Results on OOD and CAMO datasets are obtained using a model trained on CLEVRTEX only)}
\label{fig:samples}

\end{figure*}

Some segmentation prediction samples are provided in Fig \ref{fig:samples}. Other image samples are available in the supplementary material.
The main limitation of the proposed model  is the management of  shadows, which may be considered by the model as separate objects or integrated to object segmentations.

\subsection {Ablation study and additional experiments}

\newcommand{\sarif}[1]{
\hspace*{-.5em}{#1}
}

\newcommand{\starredarif}[1]{
\hspace*{-.5em}{#1}*
}

\begin{table*}[h!]
  \centering
\caption{Results of ablation study and additional experiments (results over 1 run, except for starred values, which are averages over 3 runs)}
\label{tab:ablation}
\scalebox{.8}{
\begin{tabular}{lrrrrrrrrrrrrrr}
\toprule
Dataset  &  \multicolumn{2}{c}{ \CLEVRTEX{} }& &\multicolumn{2}{c}{ \CLEVR{} }& &  \multicolumn{2}{c}{  ShapeStacks }& &  \multicolumn{2}{c}{  ObjectsRoom }

\\

\cmidrule[\lightrulewidth]{1-1}
\cmidrule[\lightrulewidth]{2-3}
\cmidrule[\lightrulewidth]{5-6}
\cmidrule[\lightrulewidth]{8-9}
\cmidrule[\lightrulewidth]{11-12}
& mIoU 
& ARI-FG 
&
&mIoU 
&  ARI-FG  
&
& mIoU 
&  ARI-FG 
&
& mIoU  
&  ARI-FG  

\\

full model AST-Seg-B3-CT (reference)

&\starredarif{79.58}&\starredarif{94.77}&&\starredarif{90.27}&\starredarif{98.26}&&\starredarif{78.84}&\starredarif{79.34}&&\starredarif{85.02}&\starredarif{87.23}\\
\cmidrule[\lightrulewidth]{1-1}
\cmidrule[\lightrulewidth]{2-3}
\cmidrule[\lightrulewidth]{5-6}
\cmidrule[\lightrulewidth]{8-9}
\cmidrule[\lightrulewidth]{11-12}

model without transformer encoder
&\sarif{75.69}&\sarif{94.41}&& \sarif{77.16}& \sarif{93.09} && \starredarif{82.99}&\starredarif{82.29}&&\starredarif{85.51}& \starredarif{88.49}\\

K = 1 + maximum number of objects 
&\starredarif{79.11}  & \starredarif{94.78} &  & \starredarif{91.03}& \starredarif{98.17}& & \sarif{78.87} & \sarif{80.05}  & &\sarif{82.90}  & \sarif{86.45} 
\\
K = 2 $\times$ maximum number of objects

&\sarif{62.10}  & \sarif{89.96} &  & \sarif{90.56}  & \sarif{98.29} & & \sarif{54.88} & \sarif{65.16}  & &\sarif{66.78}  & \sarif{78.58} 

 \\
using a Unet instead of Segformer feature generator

&\sarif{66.82}  & \sarif{88.25} &  & \sarif{90.70}  & \sarif{98.17} & & \sarif{75.51} & \sarif{77.78}  & &\sarif{85.59}  & \sarif{87.93} 
\\

random initialization of Segformer backbone

 & \sarif{61.74} & \sarif{80.22}  & &\sarif{88.94}  & \sarif{97.77}  & & \sarif{62.73} & \sarif{68.40}  & &\sarif{77.71}  & \sarif{79.23} 

\\

training without pixel entropy loss
&\sarif{70.18}  & \sarif{91.81} &  & \sarif{85.54}  & \sarif{96.09} & & \sarif{52.17} & \sarif{60.08}  & &\sarif{84.21}  & \sarif{86.19} 

\\
 training using frozen pretrained background model

 & \sarif{75.30} & \sarif{95.31}  & &\sarif{81.46}  & \sarif{98.29}  & & \sarif{55.06} & \sarif{66.24}  & &\sarif{85.82}  & \sarif{87.78} 

\\

isotropic scaling

     & \sarif{78.68} & \sarif{94.78} & & & && &  & &\sarif{84.91}  & \sarif{87.20} 

\\

anisotropic scaling

  &  & & & \sarif{87.21} & \sarif{98.53}  & &\sarif{45.47}  & \sarif{36.43} & &   &

\\

\bottomrule
\end{tabular}
}
\end{table*}
We provide in Table \ref{tab:ablation} results obtained using various ablations or modifications on the model architecture or loss function, which show that:
\begin {itemize}
 \item The model remains competitive if the transformer encoder is removed by setting $(\phi_k,  x_k,   y_k)_{1 \le k \le K} = (\phi_k^0, x_k^0,  y_k^0)_{1 \le k \le K}$. The results on the ShapeStacks and ObjectsRoom datasets are even improved with this simplified architecture, with a surprisingly strong improvement on the Shapestacks dataset, which shows the efficiency of the attention and soft-argmax mechanism. The transformer encoder is however necessary on the more complex CLEVR and CLEVRTEX datasets.
  \item Training with a number of slots slightly higher than the maximum number of objects does not lead to significant changes in the results. A more substantial increase of the number of slots however  leads to poor results on scenes with complex textures due to the increasing fragmentation the objects. This is very different from the situation observed on query-based supervised  detection models like DETR, where the number of queries has to be very high compared to the number of objects. 
 
 \item It is possible to replace the Segformer high resolution feature map generator with any other generator. The proposed model was  originally designed with a custom Unet feature map generator, which gets similar results as the Segformer model on CLEVR, ShapeStacks and ObjectsRoom, but underperforms on the more complex CLEVRTEX dataset. The architecture of this Unet is described in the supplementary material. 
 
 \item Using a pretrained backbone is necessary to get good performances with a Segformer feature map generator. 
 
 \item We  tested an alternative training scenario where the background model remains frozen during the complete training of the foreground model (125 000 iterations). The main advantage of this scenario is that it is significantly faster and requires less memory, since the backgrounds of the training images can be pre-computed and memorized. The accuracy of the results is however lower than the curriculum training scenario proposed in this paper, except for the ObjectsRoom dataset.
 
\item Switching between isotropic scaling and anisotropic scaling does not make much difference, except for the ShapeStacks dataset, where the proposed model can consider that each block tower is a single object if anisotropic scaling is enabled.

\end{itemize}

  \begin{table*}[h!]

  \caption{Training computation time with one Nvidia RTX 3090 GPU (curriculum training) }
  \centering
    \scalebox{0.8}{
        \begin{tabular}{lrrrrrrrr}
            \toprule
            \multirow{2}{*}{Dataset } &   \multirow{2}{*}{image size} &   \multicolumn{2}{c} {background model pretraining (phase 1)} & &  \multicolumn{2}{c} {full model training (phase 2 \& 3)} \\
            \cmidrule[\lightrulewidth]{3-4}
\cmidrule[\lightrulewidth]{6-7}
            &&  number of iterations & training time &  & number of iterations & training time & \\

           CLEVRTEX  & $128\times128$ &  500000& 57 h 47 mn& & 125000 & 16 h 00 mn  \\
             CLEVR  & $128\times128$ & 2500& 20 mn& & 125000 & 12 h 03 mn \\
             ObjectsRoom & $64\times64$ & 500000&  14 h 57 mn & & 125000   &  6 h 31 mn \\
             ShapeStacks & $64\times64$ &  500000&   14 h 20 mn   & &125000 & 6 h 22 mn \\ 

            \bottomrule
        \end{tabular}
  }
  \label{tab:training time}
   \vspace{-5mm}
\end{table*}
\subsection {Computation time}

All experiments have been performed using a Nvidia RTX 3090 GPU and a AMD 7402 EPYC CPU.  

Some training durations are provided in Table \ref{tab:training time}.

\section {Conclusion}
We have described in this paper a new architecture for unsupervised object-centric representation learning and object detection and segmentation, which relies on attention and soft-argmax, and shown that this new architecture substantially improves upon the state of the art on existing  benchmarks showing synthetic scenes with complex shapes and textures.  We hope this work may help to extend the scope of structured object-centric representation learning from research to practical applications.

\section{Supplementary Material}

\subsection {Hyperparameter values}

The hyperparameter values used for the proposed model are listed in Table \ref{tab:hyperparameters}.

\begin{table*}
\caption{Hyperparameter values}
  \centering
\scalebox{0.9}{

\begin{tabular}{lrrrrr}

\toprule
hyperparameter description & notation &  value \\
\midrule
Background model pretraining: \\

batch size & & 128 \\
learning rate & & $2. 10^{-3}$  \\
number of background model training iterations: \\
- datasets with fixed backgrounds (CLEVR) & & 2500 \\
- datasets with complex backgrounds (CLEVRTEX, ShapeStacks,ObjectsRoom) & & 500000 \\

\midrule
Foreground model training: \\
batch size & & 64 \\
learning rate  & & $4. 10^{-5}$  \\
Adam $\beta_1$  &  & $0.90$ \\
Adam $\beta_2$  &  & $0.98$  \\
Adam $\epsilon$ &  & $10^{-9}$  \\
number of foreground model training iterations & & 125000  \\
number of steps of phase 2 (CT scenario) &  & 30000 \\
number of steps of learning rate warmup phase &  & 5000 \\

number of steps of pixel entropy loss weight warmup phase &$N_{pixel}$ & 10000  \\

initial value of background activation before training &  $\alpha_0$  &  $e^{11}$  \\
dimension of $z_{what}$ & $d_{z_{what}}$ & 32  \\

pixel entropy loss weight & $\lambda_{pixel}$ & $1. 10^{-2}$  \\

minimum value of inverse scaling factor &$s_{min}$ & 1.3  \\
maximum value of inverse scaling factor & $s_{max}$& 24 \\

dimension of inputs and outputs of transformer encoder &$d_{T}$ & 256  \\
number of heads of transformer encoder layer & & 8 \\
dimension of feedforward transformer layer & & 512  \\
number of layers of transformer encoder & & 6  \\

\bottomrule
\end{tabular}
}
\label{tab:hyperparameters}
\end{table*}

\subsection {Pseudo-code for objects encoder and decoder}

The full encoding and rendering process is described in Algorithms \ref{alg:encoder} and \ref{alg:rendering}.
\begin{algorithm*}
\caption{Encoding}
\label{alg:encoder}
\DontPrintSemicolon

\KwIn{input image $\bX$}

\KwOut{object latents $\{\bz_{k}^{what},x_k,y_k,s_k , \alpha_k \}_{1 \ge k \ge K}$}

\tcp{feature and attention maps generation}

$(\Phi, A_1, .., A_K)$ = Segformer($\bX$)\;
\For{$k\leftarrow 1$ \KwTo $K$} 
{ $ \mathcal A_k(i,j)$  = Softmax$(A_k)(i,j) = \frac {e^{A_k(i,j)}} {\sum_{i,j} e^{A_k(i,j)}}$
 }

\tcp{computation of positions and feature vectors before transformer refinement}

\For{$i \leftarrow 1$ \KwTo $w$,  $j \leftarrow 1$ \KwTo $h$} 
 {$x(i) = 2 \frac {i-1}{w^*-1} - 1$ ; 
$ y(j) = 2 \frac {j-1}{h^*-1} - 1$ }

\For{$k\leftarrow 1$ \KwTo $K$} 
{
    $x_k^0 = \sum_{i,j}x(i) \mathcal A_k(i,j) $ ;
    $y_k^0 = \sum_{i,j}y(j )\mathcal A_k(i,j) $ \
    
    $\phi_k^0 = \sum_{i,j} \Phi(i,j) \mathcal A_k(i,j) $
     }

 \tcp{ transformer refinement of positions and feature vectors}
 
 $ ( x_k,  y_k, \phi_k)_{1 \ge k \ge K}$ = LinearProjection(TransformerEncoder(LinearEmbedding($ ( x_k^0,  y_k^0, \phi_k^0)_{1 \ge k \ge K}))) $
 
  \tcp{ latent computations}
\For{$k\leftarrow 1$ \KwTo $K$} 
{ 
$ x_k$ = clamp$(x_k, min = -1, max = 1) $ ; $ y_k$ = clamp$(y_k, min = -1, max = 1) $

$( s_k, \alpha_k, z_k^{what}) = \phi_k$ 

$ s_k = s_{min} + (s_{max}-s_{min}) \sigma(s_k) $

$ \alpha_k = e^{\alpha_k} $ }

\KwOut{ $ \{  \bz_{k}^{what},x_k,y_k,s_k , \alpha_k \}_{1 \ge k \ge K}$ }
\end{algorithm*}

\begin{algorithm*}
\caption{Rendering}
\label{alg:rendering}
\DontPrintSemicolon

\KwIn{object latents $\{\bz_{k}^{what},x_k,y_k,s_k,  \alpha_k$ \}, background image $L_0$, background mask $M_0=1$, learned background activation $\alpha_0$ or $\alpha_0(i,j)$  }

\KwOut{Image reconstruction $\hat{\bX} $}
\tcp{Obtain the object appearance $\bo_{k}$ and segmentation mask $\bm_{k}$}
\For{$k\leftarrow 1$ \KwTo $K$} 
{$\bo_{k}, \bm_{k}$ = GlimpseGenerator($\bz_{k}^\what$)\;}

\tcp{translation and scaling using a spatial transformer network (\text{STN})}
\For{$k\leftarrow 1$ \KwTo $K$} 
{
    $L_k = \text{STN}(\bo_k, x_k, y_k, s_k )$ 
    
    $M_k = \text{STN}( \bm_k,  x_k, y_k, s_k)$ }\;
    
\tcp{occlusion computations}
\For{$k\leftarrow 0$ \KwTo $K$} 
{
    $w_k = \frac {\alpha_k M_k} {\sum_{i=0}^K \alpha_i M_i} $\;
}

\tcp{combination of image layers}
$\hat{\bX} = \sum_{k = 0}^K w_k L_k $; 

\KwOut{$\hat{\bX}$}
\end{algorithm*}

\subsection {Additional implementation details}

The glimpse convolutional generator is described in Table \ref{table:glimpse}.  

\begin{table*}
 \caption{glimpse generator architecture}
  \scalebox{0.65}{
        \setlength{\tabcolsep}{2.mm}{
        
            	\begin{tabular}{cccccc}
   	 	\multicolumn{5}{c}{64x64 images}  \tabularnewline
          	 \tabularnewline
          	 \toprule
            	Layer           			& Size & Ch & Stride &Padding &  Norm./Act.   \tabularnewline \hline \hline
            	Input           			& 1       &    $d_{z_{what}}$    &     &           \tabularnewline
            	Transp Conv $2\times2$ & 2   & 64  & 2    & 0  & GroupNorm(4,64) /CELU \tabularnewline
            	Transp Conv $4\times4$ & 4  & 32    & 2   & 1   & GroupNorm(2,32)/CELU \tabularnewline
            	Transp Conv$4\times4$ & 8   & 16   & 2   & 1   & GroupNorm(1,16)/CELU \tabularnewline
            	Transp Conv $4\times4$ & 16   & 8   & 2  & 1    & GroupNorm(1,8)/CELU \tabularnewline
            	Transp Conv $4\times4$ & 32   & 4   & 2   & 1   &  \tabularnewline
            	Sigmoid & 32 & 4 \tabularnewline
            	\bottomrule
            	\end{tabular} 
	
        		\quad 
		
              	\begin{tabular}{cccccc}
    		\multicolumn{5}{c}{128x128 images}  \tabularnewline
         	 \tabularnewline
            	\toprule
            	Layer           			& Size & Ch & Stride &Padding &  Norm./Act.   \tabularnewline \hline \hline
            	Input           			& 1       &    $d_{z_{what}}$    &     &           \tabularnewline
            	Transp Conv $2\times2$ & 2   & 128  & 2    & 0  & GroupNorm(8,128) /CELU \tabularnewline
            	Transp Conv $4\times4$ & 4  & 64    & 2   & 1   & GroupNorm(4,64)/CELU \tabularnewline
            	Transp Conv$4\times4$ & 8   & 32   & 2   & 1   & GroupNorm(2,32)/CELU \tabularnewline
            	Transp Conv $4\times4$ & 16   & 16   & 2  & 1    & GroupNorm(1,16)/CELU \tabularnewline
            	Transp Conv $4\times4$ & 32   & 8   & 2   & 1   & GroupNorm(1,8)/CELU \tabularnewline
            	Transp Conv $4\times4$ & 64    & 4  & 2   & 1    \tabularnewline
		Sigmoid & 64 & 4 \tabularnewline
            	\bottomrule
            	\end{tabular}
        						}
    				}
  \label{table:glimpse}
\end{table*}

\begin{table*}
  \caption{U-net architecture (ablation study)} 
    \centering
    \scalebox{0.8}{

            \begin{tabular}{cccccc}
           
            \tabularnewline
            \toprule
            Layer           			&  Ch & Stride &Padding &  Norm./Act.   \tabularnewline \hline \hline
            Input           			&           3    &     &           \tabularnewline
            Conv $3\times3$ &     80  & 1    & 1  & BatchNorm /CELU \tabularnewline
            Downsample block &  128    &    &    & \tabularnewline
             Downsample block&  192   &    &    & \tabularnewline
            Downsample block &  256   &   &     &  \tabularnewline
            Downsample block &  256   &    &    &  \tabularnewline
            Downsample block &  256  &    &     \tabularnewline
            Center block &  256  &    &     \tabularnewline
            Upsample block &  256  &    &     \tabularnewline
            Upsample block &  256  &    &     \tabularnewline
            Upsample block &  192  &    &     \tabularnewline
            Upsample block &  128  &    &     \tabularnewline
            Upsample block &  80  &    &     \tabularnewline
            Conv $3 \times 3$ with skip connection  &   $d_\Phi$ &1&1 &   BatchNorm /CELU\tabularnewline
            Residual Conv $3 \times 3$  &  $d_\Phi$& 1 & 1 & \tabularnewline
            Conv $1 \times 1$ &   $d_\Phi$ &1&1 & \tabularnewline
            \bottomrule
            \end{tabular}
   			 }
  \label{table:unet}
\end{table*}

Synthetic datasets and preprocessing codes were downloaded from the following public repositories: 
\begin{itemize}

\item  \url{https://www.robots.ox.ac.uk/~vgg/data/clevrtex/}
\item  \url{https://ogroth.github.io/shapestacks/}
\item     \url{https://github.com/deepmind/multi_object_datasets}
\item  \url{https://github.com/applied-ai-lab/genesis}.

\end{itemize}

The Segformer pretrained weights were downloaded from the following link: 

\url{https://huggingface.co/nvidia/mit-b3}

The architecture of the U-net  implemented for the ablation study  is described in Table \ref{table:unet}. It contains a sequence of downsample blocks which output feature maps of decreasing sizes, a center block which takes as input the feature map produced by the last downsample block, and  upsample blocks, which take as input both the output of the previous upsample or center block and the feature map of the same size produced by corresponding downsample block. 
\begin {itemize}
\item  A downsample block is composed  of a convolutional layer with stride 2 and kernel size 4, with batch normalization and CELU, followed by a residual convolutional layer with stride 1 and kernel size 3 with batch normalization and CELU. 
\item  The center block is composed of a  convolutional layer with stride 1 and kernel size 3 with batch normalization and CELU. 
\item An upsample block is composed  of a residual convolutional layer with stride 1 and kernel size 3 with batch normalization and CELU, followed by a transpose convolutional layer with stride 2 and kernel size 4, with batch normalization and CELU.
\end {itemize}

\subsection{ Additional image samples}

Additional image samples  are provided in Figures 3-8.

\begin{figure*}
\centering
 \includegraphics[width=17cm]{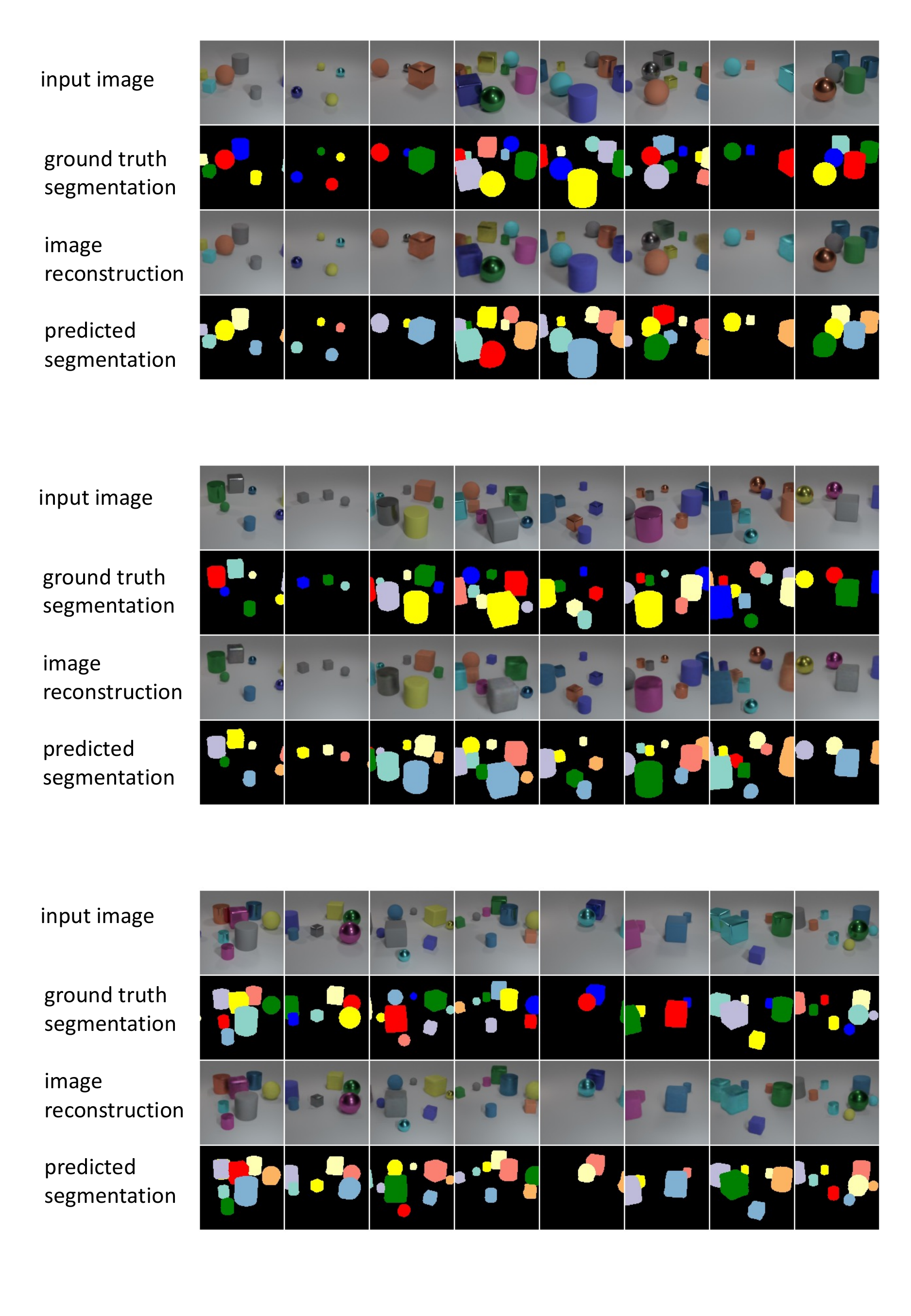} 
\caption{Examples of segmentation predictions on CLEVR test dataset}

 \vspace{-5mm}
\end{figure*}

\begin{figure*}
\centering
 \includegraphics[width=17cm]{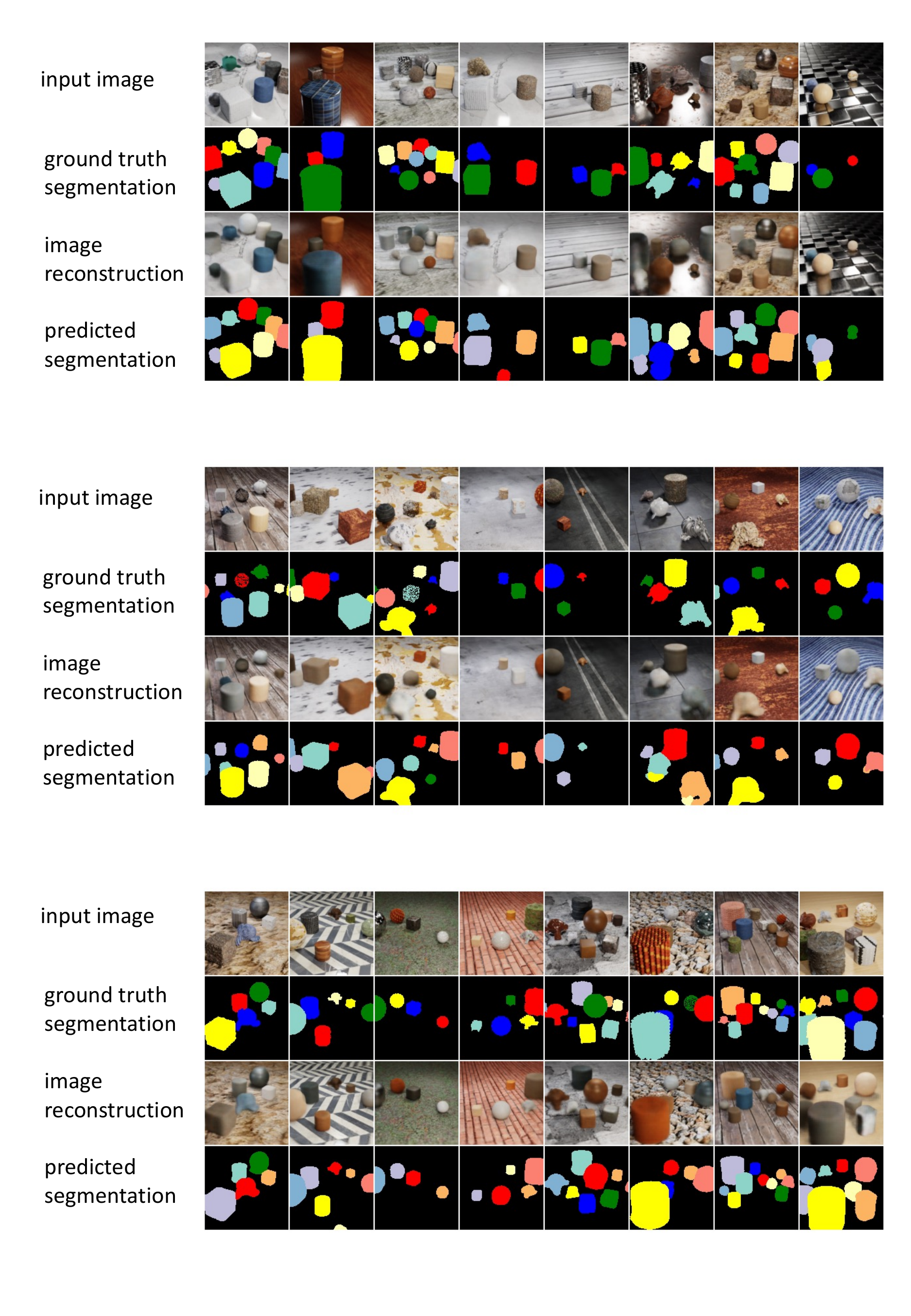} 
\caption{Examples of segmentation predictions on CLEVRTEX test dataset}

 \vspace{-5mm}
\end{figure*}

\begin{figure*}
\centering
 \includegraphics[width=17cm]{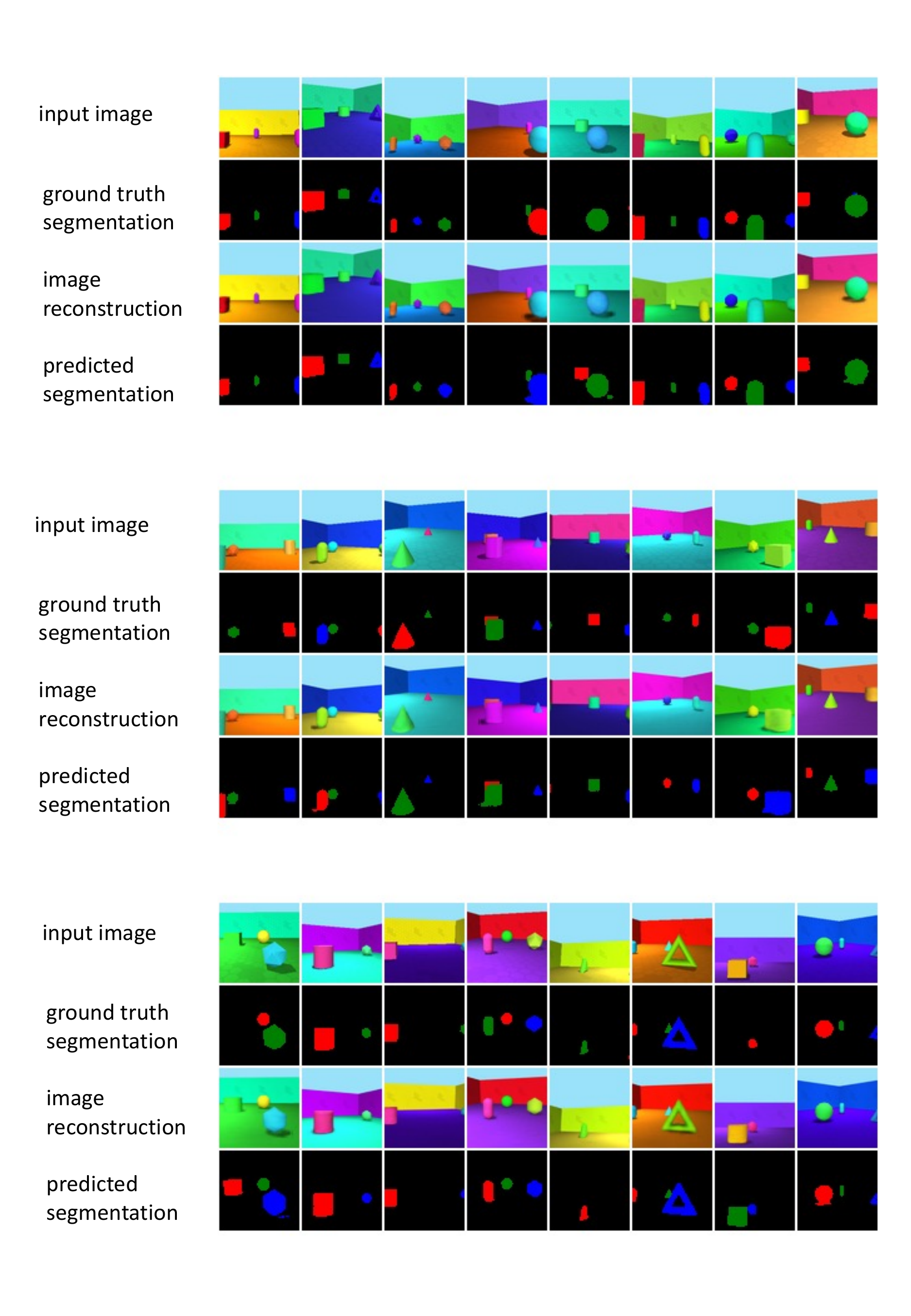} 
\caption{Examples of segmentation predictions on ObjectsRoom test dataset}

 \vspace{-5mm}
\end{figure*}

\begin{figure*}
\centering
 \includegraphics[width=17cm]{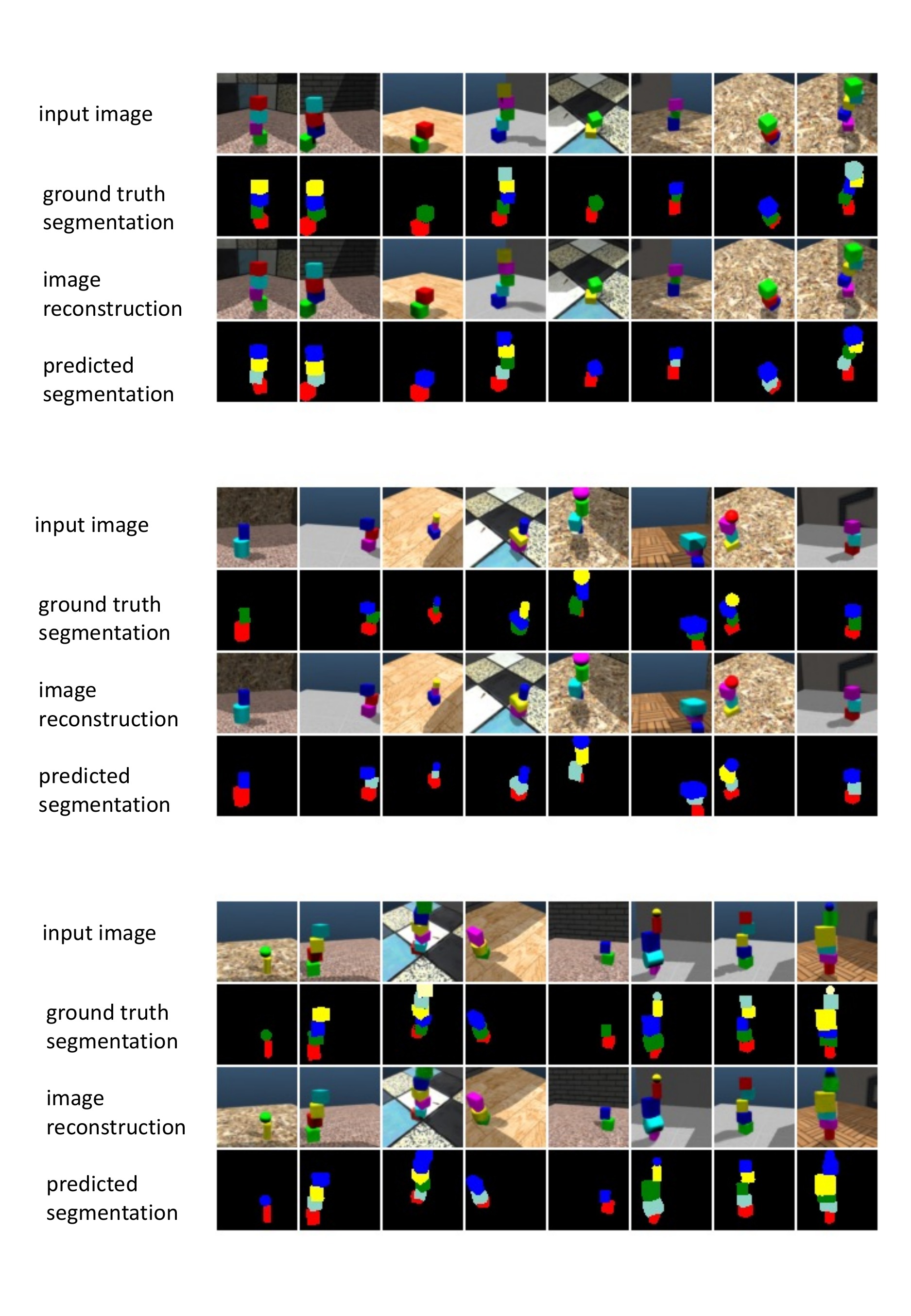} 
\caption{Examples of segmentation predictions on ShapeStacks test dataset (using a model without transformer)}

 \vspace{-5mm}
\end{figure*}

\begin{figure*}
\centering
 \includegraphics[width=17cm]{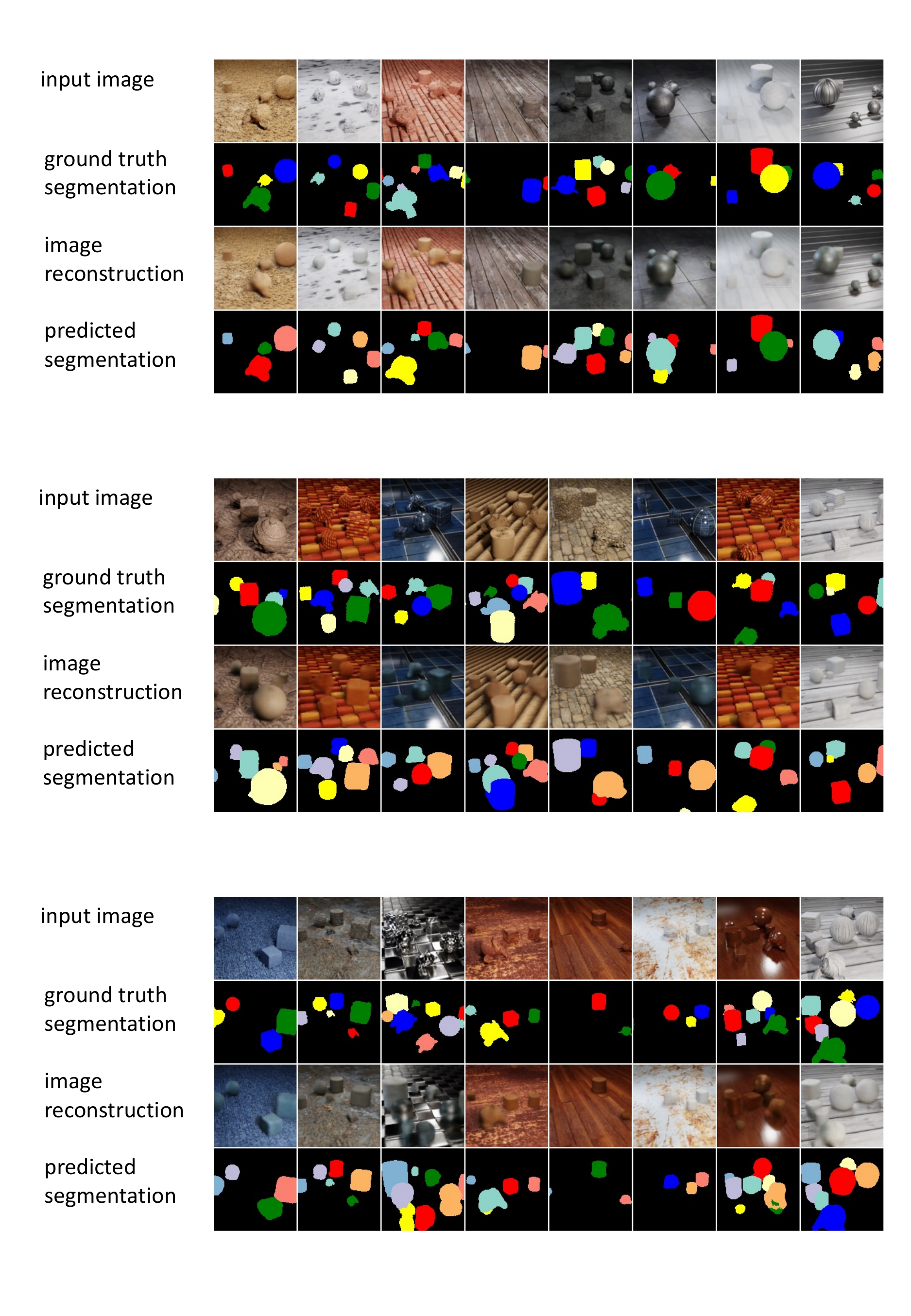} 
\caption{Examples of segmentation predictions on CAMO test dataset using a model trained on CLEVRTEX only}

 \vspace{-5mm}
\end{figure*}

\begin{figure*}
\centering
 \includegraphics[width=17cm]{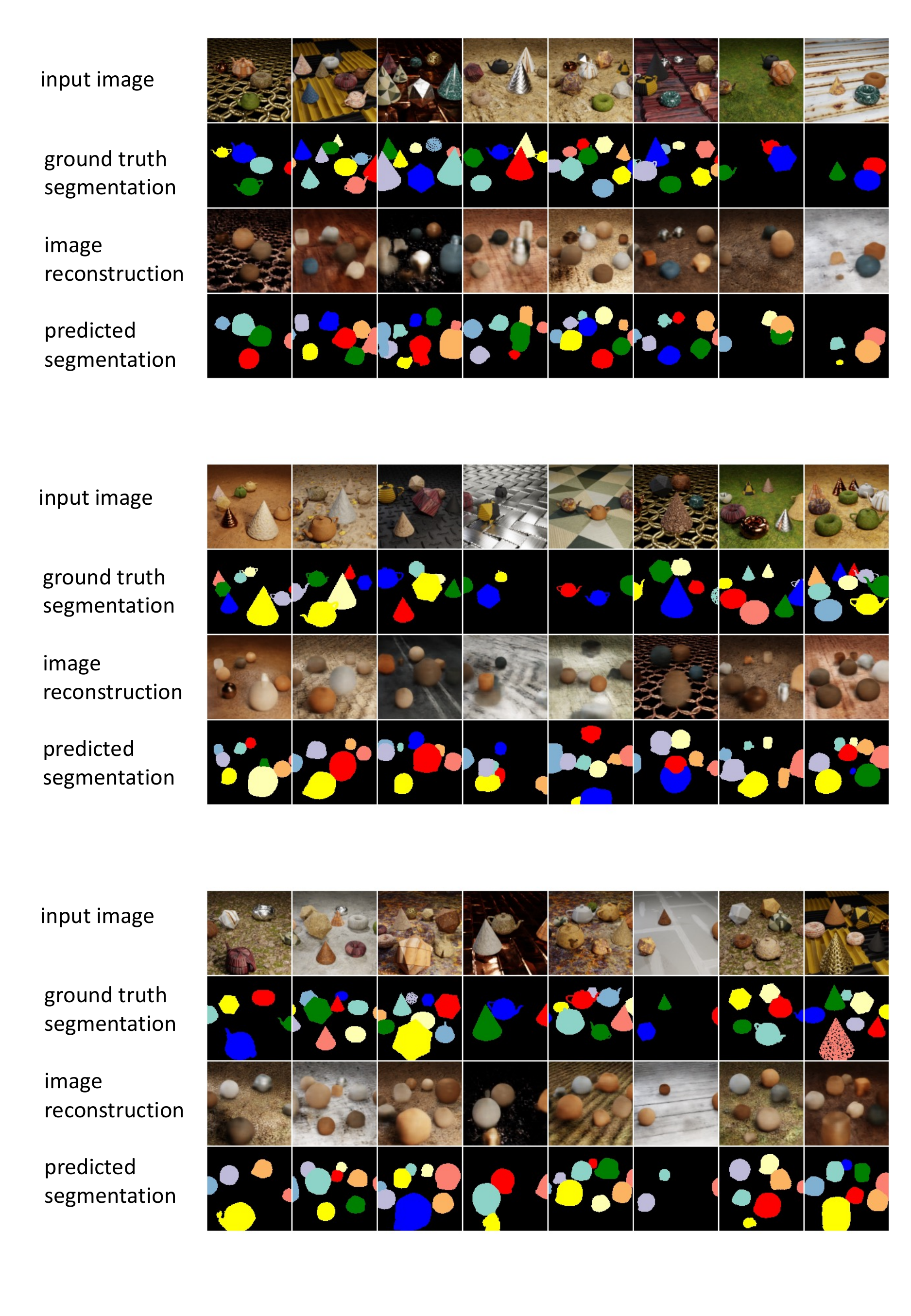} 
\caption{Examples of segmentation predictions on OOD test dataset using a model trained on CLEVRTEX only}

 \vspace{-5mm}
\end{figure*}

\newpage

{\small
\bibliographystyle{ieee_fullname}
\bibliography{AST2}
}

\end{document}